\documentclass[journal,twocolumn]{IEEEtran} 
\usepackage[utf8]{inputenc}

\usepackage{amsmath,amssymb,amsthm}
\usepackage{mathptmx} 

\usepackage{graphicx}
\graphicspath{{figures/}}  
\usepackage{epstopdf}
\usepackage{pgfplots}
\pgfplotsset{compat=1.18}
\usepackage{subcaption}
\usepackage{float}
\usepackage{changepage} 
\usepackage{placeins}

\usepackage{booktabs}
\usepackage{tabularx,colortbl,multirow,array}

\usepackage{algorithm}
\usepackage{algpseudocode}

\usetikzlibrary{positioning, arrows.meta, shapes.geometric, arrows}

\usepackage{subcaption} 

\usepackage[compact]{titlesec}  
\titlespacing*{\section}{0pt}{*1.3}{*0.8}
\titlespacing*{\subsection}{0pt}{*1.2}{*0.68}  

\usepackage{bm}

\usepackage{caption}
\usepackage{hyperref}
\usepackage{enumitem}
\usepackage{tcolorbox}
\tcbuselibrary{fitting}
\usepackage{mdframed}
\usepackage[hang,flushmargin]{footmisc}
\usepackage{cite}
\usepackage{url}
\usepackage{xcolor}
\usepackage{amsthm}  
\usepackage{microtype}
\usepackage{etoolbox}
\usepackage{indentfirst} 
\usepackage{geometry}
\usepackage{tikz}
\usetikzlibrary{shapes.geometric, arrows.meta, positioning, calc}

\title{Few Shot Semi-Supervised Learning for Abnormal Stop Detection from Sparse GPS Trajectories}

\author{
    Muhammad Ayub Sabir,
    Junbiao Pang,
    Jiaqi Wu,
    Fatima Ashraf%
    \thanks{Email: sabir@emails.bjut.edu.cn}%
    \thanks{Email: junbiao\_pang@bjut.edu.cn}%
    \thanks{Email: qi2019kb@163.com}%
    \thanks{Email: fatimaashraf@emails.bjut.edu.cn}
}

\begin{document}

\maketitle

\begin{abstract}

Abnormal stop detection (ASD) in intercity coach transportation is critical for ensuring passenger safety, operational reliability, and regulatory compliance. However, two key challenges hinder ASD effectiveness: sparse GPS trajectories, which obscure short or unauthorized stops, and limited labeled data, which restricts supervised learning. Existing methods often assume dense sampling or regular movement patterns, limiting their applicability. To address data sparsity, we propose a Sparsity-Aware Segmentation (SAS) method that adaptively defines segment boundaries based on local spatial-temporal density. Building upon these segments, we introduce three domain-specific indicators to capture abnormal stop behaviors. To further mitigate the impact of sparsity, we develop Locally Temporal-Indicator Guided Adjustment (LTIGA), which smooths these indicators via local similarity graphs. To overcome label scarcity, we construct a spatial-temporal graph where each segment is a node with LTIGA-refined features. We apply label propagation to expand weak supervision across the graph, followed by a GCN to learn relational patterns. A final self-training module incorporates high-confidence pseudo-labels to iteratively improve predictions. Experiments on real-world coach data show an AUC of 0.854 and AP of 0.866 using only 10 labeled instances, outperforming prior methods. The code and dataset are publicly available at \href{https://github.com/pangjunbiao/Abnormal-Stop-Detection-SSL.git}{https://github.com/pangjunbiao/Abnormal-Stop-Detection-SSL.git}.

\end{abstract}

\begin{IEEEkeywords}
Abnormal Stop Detection, Sparse GPS Trajectories, Semi-Supervised Learning, Graph Convolutional Networks (GCN)
\end{IEEEkeywords}

\section{Introduction}

Public transportation plays a key role in enhancing mobility, supporting economic growth, and promoting sustainability~\cite{oh2024enhancing, fan2019research, zhang2014understanding}. Long-distance coaches, in particular, offer cost-effective and energy-efficient alternatives to rail and air travel, especially in regions with limited infrastructure~\cite{de2024banning, nature2024longdistance, chen2013iboat}. These services connect remote areas to urban centers, fostering accessibility and socioeconomic development~\cite{stokenberga2025connecting, oh2024enhancing, van2020preferences}. However, managing coach operations is challenging due to extended travel durations and variable traffic patterns~\cite{deng2024unsupervised, pang2017discovering, xu2024perimeter}.

A core challenge in abnormal stop detection (ASD) lies in identifying deviations from planned stop locations, such as unauthorized pick-ups, which pose safety and regulatory risks~\cite{ceccato2022crime, zhang2017correction}. The difficulty of detecting such deviations is further exacerbated by low-frequency GPS trajectories, where large sampling intervals ($\Delta t$) reduce temporal granularity and obscure unauthorized stops~\cite{deng2024unsupervised, chen2024micro}. Moreover, labeled data is extremely limited, as only a small fraction of GPS points are manually annotated~\cite{berte2024enhancing}.

To address the challenges of sparse GPS data and label scarcity, we propose a four-stage pipeline. First, Sparsity-Aware Segmentation (SAS) adaptively partitions trajectories into locally dense segments to preserve behavioral granularity under sparse sampling.  Second, we define three domain-specific indicators—\textit{Temporal Influence Score (TIS)}, \textit{Maximum Speed Deviation (MSD)}, and \textit{Top-$k$ Aggregated Temporal Score (TTA@k)}—designed to capture abnormal stop behaviors. 
Third, to improve indicator reliability and mitigate the impact of sparsity, we introduce Locally Temporal-Indicator Guided Adjustment (LTIGA), which smooths features via similarity-based temporal graphs, refining signal quality without relying on rigid assumptions.
Finally, to address limited labeled data, we construct a spatial-temporal graph where each segment is a node with LTIGA-refined features. We apply label propagation to diffuse weak supervision across the graph. A self-training module further enhances prediction by iteratively refining labels using high-confidence pseudo-labels, improving robustness under label scarcity. Overall, this semi-supervised strategy enables robust learning under sparse, noisy GPS trajectories—limitations that prior methods fail to overcome. This paper makes the following key contributions:
\begin{itemize}[left=0pt, label=\textbullet]
    \item We propose \textbf{Sparsity-Aware Segmentation (SAS)}, a novel trajectory partitioning method that adaptively segments GPS traces based on local spatial-temporal density, effectively preserving behavioral granularity under sparse sampling.
    
    \item We design \textbf{three interpretable, domain-specific indicators}—TIS, MSD, and TTA@k—to detect abnormal stop behaviors at the segment level, enhancing robustness against data sparsity and noise.
    
    \item We develop \textbf{Locally Temporal-Indicator Guided Adjustment (LTIGA)}, which refines indicator quality by smoothing them over local temporal similarity graphs, improving feature consistency without rigid assumptions on speed or timing.
    
    \item We construct a \textbf{semi-supervised learning framework} that integrates label propagation, Graph GCNs, and self-training, enabling accurate ASD with minimal labeled data under real-world GPS sparsity.
\end{itemize}

\section{Related Work}

\subsection{\textbf{Abnormal Behavior Detection in ITS}}
Anomaly detection in intelligent transportation systems (ITS) has been widely explored to identify deviations in driving behavior, safety-critical events, and non-compliant operational patterns~\cite{pang2024finding, pang2018learning, kumar2017visual, chen2011real}. Many studies leverage spatiotemporal data to support these tasks. For instance, Boateng et al.~\cite{boateng2023gps} and Park et al.~\cite{park2024unsupervised} utilized GPS analytics with handcrafted or unsupervised features to detect unsafe driving. Yu and Huang~\cite{yu2022deep} and Li et al.~\cite{li2023temporal} developed deep encoder-decoder and temporal attention models to capture sequential trajectory anomalies. Zhang et al.~\cite{zhang2023semi} proposed a semi-supervised method using surrogate safety metrics, while Yang et al.~\cite{yang2021driving} conducted real-time anomaly detection on GNSS traces from public buses. Kumar et al.~\cite{kumar2017visual} explored visual-numeric fusion for outlier detection in vehicle trajectories.

These studies demonstrate the effectiveness of spatiotemporal data for abnormal behavior detection. However, most existing approaches are designed for high-frequency GPS scenarios—typically 1 Hz or higher—where fine-grained motion patterns can be accurately captured~\cite{yu2022deep, li2023temporal}. In contrast, our work addresses long-distance coach networks, where GPS trajectories are sparsely sampled at intervals of 30–60 seconds~\cite{deng2024unsupervised}, posing unique challenges for conventional detection methods.

\subsection{\textbf{Stop Inference from Sparse GPS Trajectories}}

Inferring reliable stops from sparse GPS trajectories is challenging due to low sampling rates, signal noise, and limited contextual information. To address this, Deng et al.~\cite{deng2024unsupervised} proposed a linear speed approximation model tailored to low-frequency GPS data in long-distance coach services. Bert\`e et al.~\cite{berte2024enhancing} modeled routine behaviors while incorporating neighborhood-level spatial continuity. Map-matching techniques have also improved inference accuracy: Ozdemir and Topcu~\cite{ozdemir2018hybrid} applied a hybrid hidden Markov model (HMM), and Zhang et al.~\cite{zhang2015shortest} combined shortest-path constraints with vehicle traces to correct GPS deviations. Chen et al.~\cite{chen2017road} introduced a collaborative path inference framework to reconstruct complete trajectories from sparse GPS snippets, enhancing stop detection. 

While these methods improve stop inference under sparse GPS conditions, they typically assume either fully supervised settings or rely on static heuristics. This motivates the design of flexible, sparsity-aware models that can generalize across varying sampling intervals and real-world trajectory irregularities

\subsection{\textbf{Graph-Based Semi-Supervised Learning for Anomaly Detection}}

Recent advancements extend Graph-based semi-supervised learning (GSSL) in complex and evolving systems. Zheng et al.~\cite{zheng2024generative} proposed a generative GSSL model that fuses representations from labeled and unlabeled nodes to improve anomaly identification. Song et al.~\cite{song2024novel} introduced a graph structure learning framework tailored to dynamic IoT environments, achieving robustness under irregular sampling.  For evolving graph scenarios, Chen et al.~\cite{chen2025semi} developed EL$^{2}$-DGAD, which combines transformer-based encoders with ego-context contrastive learning under extreme label scarcity. Tian et al.~\cite{tian2023sad} proposed SAD, a dynamic GSSL approach using pseudo-label contrastive learning and memory modules. Latif-Martínez et al.~\cite{latif2023detecting} applied graph neural networks for contextual anomaly detection in streaming data, while Zheng et al.~\cite{zheng2023correlation} formulated a correlation-aware spatiotemporal graph model for multivariate time-series anomalies.

Despite their success in domains like IoT and time-series analytics, existing GSSL methods lack adaptations for transportation scenarios involving sparse, segment-based GPS data. Addressing this gap requires trajectory-aware graph models that operate at the segment level, incorporate temporal continuity, and remain robust under extreme label scarcity.

\section{Methodology}

We consider GPS trajectory data from long-distance coach trips, where each GPS data point comprises longitude (\( \text{lng}_i \)), latitude (\( \text{lat}_i \)), timestamp (\( t_i \)) indicating the time of recording, instantaneous velocity (\( v_i \)), and engine state (\( f_i \)), which denotes whether the coach is moving or stationary. Stay time ($s_i$) at each GPS point $p_i$ is estimated as the time interval during which the coach maintains a stationary engine state (i.e., $f_i = 0$) and velocity near zero ($v_i \approx 0$), aggregated over consecutive GPS samples. A summary of the dataset is provided in Section~\ref{sec:dataset_analysis}-Table~\ref{tab:dataset_summary}.

\newtheorem*{definition}{Coach Trip}
\textbf{Coach Trip:} A coach trip \( T \) is defined as an ordered sequence of GPS data points:
\begin{equation}
T = \{p_1, p_2, \dots, p_N\}, \quad \text{with } 0 < t_{i+1} - t_i \leq \Delta t
\end{equation}
where \( p_i = (\text{lng}_i, \text{lat}_i, t_i, v_i, f_i) \) and \( \Delta t \) denotes the maximum sampling interval, reflecting the low-frequency nature of the data.

\subsection{Normal and Abnormal Stop Classification}

Due to sparse GPS sampling and signal noise, stop classification relies on derived features such as velocity and stay time. We define four representative cases for identifying normal and abnormal stop behavior.

\begin{itemize}
    \item \textbf{Case 1: Explicit Stop Detection} — A stop is detected at point \( p_i \) if both zero velocity and nonzero stay time are observed:
    \begin{equation}
        v_m = 0 \text{ and } s_i > 0 \Rightarrow \text{Stop at } p_i
    \end{equation}

    \item \textbf{Case 2: Velocity-Based Estimation} — When \( v_m \) or \( s_i \) is unreliable, the velocity is estimated between two points as:
    \begin{equation}
        v_i \approx \frac{2d}{t_{i+1} - t_i} + v_m
    \end{equation}
    where \( d \) is the spatial distance between \( p_i \) and \( p_{i+1} \), and \( v_m \) denotes a small velocity margin used to account for GPS noise and under-sampling effects. It is set to zero when detecting explicit stops (Equ 2), and used as a correction term in velocity estimation under sparse conditions (Equ 3) to improve robustness. A stop is inferred if:
    \begin{equation}
        v_i = 0 \text{ and } d \leq d_{\text{threshold}} \Rightarrow \text{Stop at } p_i
    \end{equation}
\end{itemize}

These two cases detect normal stop behavior. Abnormal stops, by contrast, are defined as deviations from expected patterns in location or duration. Two key cases are:

\begin{itemize}
    \item \textbf{Case 3: Temporal Deviation} — A stop is considered abnormal if its duration significantly exceeds the expected average for that route:
    \begin{equation}
        s_i \gg \bar{s}_{\text{route}} \Rightarrow \text{Abnormal stop at } p_i
    \end{equation}
    where \( \bar{s}_{\text{route}} \) is the average stay time for the route, computed from historical normal stops.

    \item \textbf{Case 4: Low-Speed Short Pauses} — A stop is also considered abnormal if the coach moves slowly with a minimal stay duration:
    \begin{equation}
        0 < v_i \leq v_{\text{low}}, \quad 0 < s_i \leq s_{\text{min}} \Rightarrow \text{Abnormal stop at } p_i
    \end{equation}
    where \( v_{\text{low}} \) is a low-speed threshold (e.g., 5 km/h), and \( s_{\text{min}} \) is the minimum meaningful stop duration, such as brief pauses under 40 seconds.
\end{itemize}

These four cases form the decision logic for detecting both normal and abnormal stops under sparse GPS conditions. The classification logic is operationalized in Algorithm~\ref{alg:stop_and_abnormal_detection}.

\begin{algorithm}[H]
\caption{Normal and Abnormal Stop Detection from Sparse GPS}
\label{alg:stop_and_abnormal_detection}
\begin{algorithmic}[1]
\Require Consecutive GPS points $p_i$ and $p_{i+1}$; maximum sampling interval $\Delta t$; spatial threshold $d_{\text{threshold}}$; low-speed threshold $v_{\text{low}}$; stay time threshold $s_{\text{min}}$; small velocity tolerance $\epsilon$; historical average stop duration $\bar{s}_{\text{route}}$
\Ensure Classification of $p_i$ as Normal Stop, Abnormal Stop, or No Stop

\State Compute spatial distance $d \leftarrow \text{Haversine}(p_i, p_{i+1})$
\State Compute time gap $\Delta t \leftarrow t_{i+1} - t_i$
\State Estimate velocity $v_i \leftarrow \frac{2d}{\Delta t} + v_m$

\If{$v_m = 0$ \textbf{and} $s_i > 0$}
    \State \textbf{Stop Detected}
\ElsIf{$v_i \leq \epsilon$ \textbf{and} $d \leq d_{\text{threshold}}$}
    \State \textbf{Stop Detected}
\Else
    \State \textbf{No Stop Detected}
\EndIf

\If{Stop Detected}
    \If{$s_i \gg \bar{s}_{\text{route}}$}
        \State \textbf{Abnormal Stop (Extended Duration)}
    \ElsIf{$0 < v_i \leq v_{\text{low}}$ \textbf{and} $0 < s_i \leq s_{\text{min}}$}
        \State \textbf{Abnormal Stop (Low-Speed Pause)}
    \Else
        \State \textbf{Normal Stop}
    \EndIf
\EndIf
\end{algorithmic}
\end{algorithm}

\paragraph{Challenges with Sparse and Noisy GPS Data}

Reliable stop detection depends on derived signals such as velocity, stay time, and spatial displacement. However, under real-world conditions, these indicators degrade due to sparse sampling and GPS noise. When the interval between consecutive GPS points (\(t_{i+1} - t_i\)) significantly exceeds the actual stop duration (\(t_{\text{stop}}\)), short-duration stops may be entirely missed. Additionally, noisy or incomplete measurements can distort feature estimates, making it difficult to distinguish brief movement from legitimate stops. These limitations contribute to both false positives and missed detections in stop classification.

\subsection{\textbf{Sparsity-Aware Segmentation (SAS)}}

Effective ASD requires partitioning GPS trajectories into coherent spatial-temporal segments, particularly under sparse and irregular sampling. To address this, we propose \textbf{SAS}, a dynamic method that adaptively determines segment boundaries based on local spatial and temporal density. Unlike conventional fixed-threshold techniques~\cite{deng2024unsupervised, etemad2019trajectory, guo2018gps}, SAS computes trajectory-specific thresholds that reflect real movement variations rather than arbitrary cutoffs.

Given two consecutive GPS points \(p_i\) and \(p_j\), a new segment is created if the spatial distance \(D_{ij}\) or temporal gap \(\Delta t_{ij}\) exceeds an adaptive threshold $\lambda_d, \lambda_t$:

\begin{equation}
\text{SegmentBreak}_{ij} = 
\begin{cases}
1, & \text{if } D_{ij} > \lambda_d \text{ or } \Delta t_{ij} > \lambda_t \\
0, & \text{otherwise}
\end{cases}
\end{equation}

\begin{figure}[tb]
  \centering
  \includegraphics[width=0.93\linewidth]{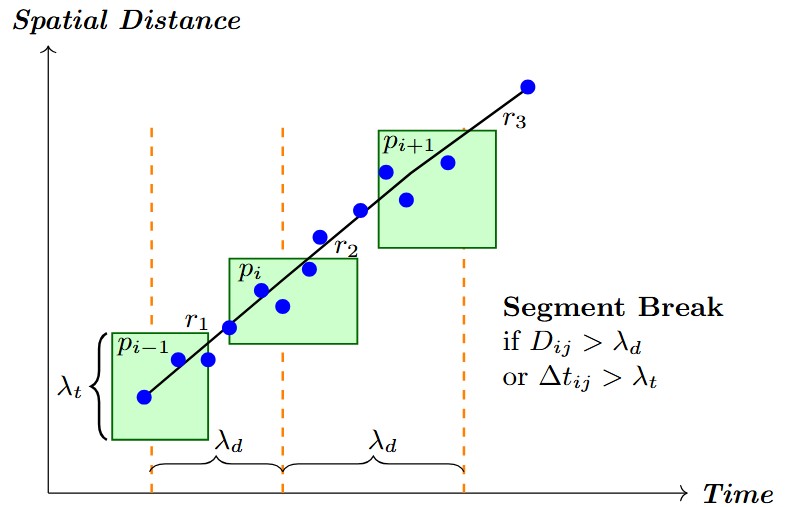}
  \caption{Illustration of Sparsity-Aware Segmentation (SAS). A segment break is introduced when the spatial distance \(D_{ij}\) or time gap \(\Delta t_{ij}\) between consecutive GPS points exceeds adaptive thresholds \(\lambda_d\) or \(\lambda_t\), respectively. These thresholds are dynamically computed using Equation~\ref{eq:adaptive_thresholds}.}
  \label{fig:sas_diagram}
\end{figure}

Figure~\ref{fig:sas_diagram} illustrates this segmentation logic. When either distance or time exceeds its respective threshold, a segment break is introduced. The thresholds \(\lambda_d\) (distance) and \(\lambda_t\) (time) are computed adaptively using the mean and standard deviation of all pairwise distances and time gaps in a trip:

\begin{equation}
\lambda_d = \mu_D + \alpha \cdot \sigma_D, \quad \lambda_t = \mu_T + \beta \cdot \sigma_T
\label{eq:adaptive_thresholds}
\end{equation}

Here, \( \mu_d \) and \( \mu_t \) denote the mean spatial distance and mean temporal gap between consecutive GPS points in a given trip, while \( \sigma_d \), \( \sigma_t \) represent their standard deviations. These statistics are computed independently for each trajectory, allowing the thresholds to adapt to intra-trip sparsity patterns. The hyperparameters \( \alpha \) and \( \beta \) control the sensitivity of segment breaking and were empirically selected based on validation performance to balance over- and under-segmentation, ensuring robustness across varying trip sparsity levels.

The trajectory is then segmented into \(R = \{r_1, r_2, \dots, r_M\}\), where each segment \(r_m\) consists of consecutive GPS points satisfying:

\begin{equation}
d(p_{i+1}, p_i) \leq \lambda_d \quad \text{and} \quad t_{i+1} - t_i \leq \lambda_t
\label{eq:sas_constraints}
\end{equation}

When either condition is violated, the current segment ends, and a new one begins. This ensures each segment preserves local behavioral coherence while minimizing over-segmentation.

\subsection{\textbf{Designing Indicators for ASD}}

We design three interpretable indicators—TIS, MSD, and TTA@k—to quantify temporal deviations, speed irregularities, and dominant stop behaviors under sparse GPS conditions. 
\noindent\textbf{Definition 1: Temporal Influence Score (TIS)} \textit{TIS quantifies how strongly a stop deviates from typical duration patterns within its segment.}
\begin{equation}
\text{TIS}_{ij} = \frac{s_{ij} - \tilde{s}_i}{\sigma_i + \epsilon},
\end{equation}

where \(s_{ij}\) is the stay time of the \(j\)-th stop in segment \(r_i\), \(\tilde{s}_i\) is the median stay time in \(r_i\), \(\sigma_i\) is the standard deviation, and \(\epsilon\) is a small constant for numerical stability. TIS is clipped to the range \([0, 3]\) to suppress extreme outliers. Higher TIS values highlight stops with unusual durations compared to the segment norm.

\noindent\textbf{Definition 2: Maximum Speed Deviation (MSD)} \textit{MSD captures speed irregularity within a segment.}
\begin{equation}
\text{MSD}_i = \left(v_{\text{max},i} - v_{\text{mean},i}\right) + \delta \cdot v_{\text{mean},i}
\end{equation}

where, \(v_{\text{max},i}\) and \(v_{\text{mean},i}\) are the maximum and mean speeds in segment \(r_i\), and \(\delta\) is a tunable hyperparameter that adjusts sensitivity. High MSD values indicate sudden accelerations or decelerations, often linked to unplanned or abrupt stops.

\noindent\textbf{Definition 3: Top-$k$ Aggregated Temporal Score (TTA@k)} \textit{TTA@k emphasizes dominant stop durations via soft aggregation.}
\begin{equation}
\text{TTA@k}_i = \sum_{j=1}^{k} \alpha_j \cdot \tilde{s}_{ij}, \quad \text{where} \quad \alpha_j =  \frac{e^{\tilde{s}_{ij}}}{\sum_{l=1}^{k} e^{\tilde{s}_{il}}}
\end{equation}
\(\tilde{s}_{ij}\) is the normalized duration of the \(j\)-th longest stop in segment \(r_i\), and \(\alpha_j\) is its softmax-based weight. This formulation prioritizes high-impact durations while remaining robust to varying segment lengths.

\noindent\textbf{Confidence Score.} \textit{Each segment’s indicators are weighted by a confidence score based on data density and motion stability.}
\begin{equation}
\text{Confidence}_i = \frac{N_i}{N_{\text{max}}} \times \frac{V_{\text{avg},i}}{V_{\text{max},i} + \epsilon}
\end{equation}
Here, \(N_i\) is the number of GPS points in segment \(r_i\), \(V_{\text{avg},i}\) is the average speed, and \(V_{\text{max},i}\) is the maximum speed. A higher confidence score indicates more reliable signals for anomaly detection.

These indicators are particularly critical under sparse GPS conditions, where traditional methods struggle due to missing or irregular measurements. They provide robust intermediate representations that capture key behavioral anomalies, serving as inputs to the subsequent LTIGA module for indicator refinement.

\subsection{\textbf{Locally Temporal-Indicator Guided Adjustment (LTIGA)}}\label{sec:ltiga}

We propose \textbf{LTIGA} to enhances indicator reliability by leveraging intra-segment similarity among GPS points.
LTIGA begins by constructing a local similarity graph within each segment $r_i$. Each GPS point is represented by an indicator vector $\mathbf{x}_i = (\text{TIS}_i, \text{MSD}_i, \text{TTA@k}_i) \in \mathbb{R}^3$. To ensure scale invariance and comparability, the indicator vectors are standardized $ 
\tilde{\mathbf{x}}_i = \frac{\mathbf{x}_i - \boldsymbol{\mu}_k}{\boldsymbol{\sigma}_k + \epsilon}
$.

Pairwise similarities between points are computed using cosine similarity, refined by a Gaussian kernel to construct the weighted similarity matrix:

\begin{equation}
\mathbf{W}_{ij} = \exp\left( -\frac{(1 - \cos(\tilde{\mathbf{x}}_i, \tilde{\mathbf{x}}_j))^2}{2\sigma^2} \right)
\end{equation}

where, \(\sigma\) controls the smoothness of influence—lower values make the kernel more sensitive to small variations, while higher values allow broader smoothing.

To perform adjustment, each indicator vector \(\tilde{\mathbf{x}}_i\) is smoothed using a weighted average over its top-\(k\) most similar neighbors:

\begin{equation}
\tilde{\mathbf{x}}_i' = \frac{1}{Z_i} \sum_{j \in N_k(i)} W_{ij} \cdot \tilde{\mathbf{x}}_j,
\end{equation}

where \(N_k(i)\) denotes the \(k\)-nearest neighbors and \(Z_i = \sum_{j \in N_k(i)} W_{ij}\) is the normalization factor.  This smoothing process anchors outlier points toward the general behavior of the segment, thereby improving the quality of the features.

Finally, smoothed vectors are transformed back to their original scale,
\begin{equation}
\mathbf{x}_i^{\text{smooth}} = \tilde{\mathbf{x}}_i' \cdot \boldsymbol{\sigma}_k + \boldsymbol{\mu}_k
\end{equation}

The resulting indicator vectors are more robust to sparsity and noise, providing reliable representations of segment behavior under irregular sampling. 

\subsection{\textbf{Graph Construction and Label Propagation}}\label{sec:weighted-ltiga}

To address the challenge of limited labeled data under sparse GPS sampling, we leverage structural dependencies among trajectory segments and propagates label information through a spatial-temporal graph.

\textbf{Graph Definition.} We construct a global spatial-temporal graph \( G = (V, E) \), where each node \( v_i \in V \) represents a trajectory segment \( r_i \) produced by SAS. Each node is associated with a feature vector derived from LTIGA-smoothed indicators, further scaled by the segment-level confidence score to reflect data reliability:

\begin{equation}
\mathbf{x}_i^{\text{w}} = C_i \cdot \mathbf{x}_i^{\text{smooth}} = [\text{TIS}_i^{\text{w}}, \text{MSD}_i^{\text{w}}, \text{TTA@k}_i^{\text{w}}] \in \mathbb{R}^3
\end{equation}

Here, \( \mathbf{x}_i^{\text{smooth}} \) denotes the refined indicator vector from LTIGA (Section~\ref{sec:ltiga}), and \( C_i \) is the segment-level confidence score. This confidence-weighted vector ensures that each node representation encodes both reliable movement behavior and data quality, improving robustness in subsequent classification.

\textbf{Edge Construction.} Edges in the graph encode two types of spatial-temporal relationships to capture both local segment structure and global behavioral similarity:

\begin{itemize}
    \item \textbf{Intra-Segment Edges:} To preserve local continuity, we connect consecutive GPS points within the same segment \( r_i \) based on their temporal order as follows:
    \begin{equation}
    e_{ij} = 1, \quad \text{if } v_i, v_j \in r_i \text{ and } t_j = t_i + 1
    \end{equation}

    \item \textbf{Inter-Segment Edges:} To capture broader behavioral similarities, the edge weight uses cosine similarity between their confidence-weighted indicator vectors:
    \begin{equation}
    w_{ij} = \cos(\mathbf{x}_i^{\text{w}}, \mathbf{x}_j^{\text{w}}), \quad \text{if } |t_i - t_j| \leq \Delta t_{\text{max}}
    \end{equation}
    Here, \( \Delta t_{\text{max}} \) is a tunable temporal window controlling the neighborhood scope. This hybrid edge design integrates fine-grained local structure and global behavioral alignment.
\end{itemize}

\textbf{Label Propagation.} To infer labels for unlabeled trajectory segments, we apply label propagation over the constructed graph. A similarity matrix \( W \) is computed using an RBF kernel $w_{ij} = \exp\left( -\frac{\|\mathbf{x}_i^{\text{w}} - \mathbf{x}_j^{\text{w}}\|^2}{2\sigma^2} \right) $.

Let \( F \in \mathbb{R}^{|V| \times C} \) denote the soft label matrix over all nodes, where \( C \) is the number of classes (e.g., normal and abnormal stops). Label propagation is formulated as minimizing the following energy functional:

\begin{equation}
Q(F) = \sum_{i,j} w_{ij} \| F_i - F_j \|^2
\end{equation}

This formulation encourages label consistency across similar nodes. To mitigate error propagation under extreme label sparsity, we employ a high-confidence pseudo-labeling strategy. A node receives a pseudo-label only if its predicted class probability surpasses a strict threshold (e.g., \( \geq 0.995 \) for abnormal, \( \leq 0.005 \) for normal). This ensures that only highly certain predictions contribute to training. Additionally, we enforce class-balance constraints and limit the number of accepted pseudo-labels per iteration to maintain diversity and prevent drift in the label distribution.

\subsection{\textbf{Graph Convolutional Network (GCN) and Self-Training}}

We apply a GCN over the global graph constructed from LTIGA-refined and confidence-weighted segment features.

Let each node \(v_i \in V\) be initialized with the feature vector \(h_i^{(0)} = \mathbf{x}_i^{\text{w}}\), where \(\mathbf{x}_i^{\text{w}}\) denotes the LTIGA-smoothed and confidence-weighted indicator vector defined in Section~\ref{sec:weighted-ltiga}. The GCN updates node representations layer-wise as:
\begin{equation}
h_i^{(l+1)} = \sigma \left( \sum_{j \in N(i)} \frac{1}{c_{ij}} W^{(l)} h_j^{(l)} + b^{(l)} \right)
\end{equation}
where \(W^{(l)}\) and \(b^{(l)}\) are the trainable weight and bias at layer \(l\), \(c_{ij} = \sqrt{\deg(i)\deg(j)}\) is a symmetric normalization factor, and \(\sigma(\cdot)\) is a nonlinear activation function such as ReLU.

After \(L\) layers, final class probabilities are predicted as:
\begin{equation}
\hat{y}_i = \text{softmax}(h_i^{(L)})
\end{equation}

\paragraph{Self-Training Module.}
To further improve generalization under limited labeled data, we adopt a confidence-aware self-training mechanism. The model selects high-confidence predictions as pseudo-labels and retrains using both original and pseudo-labeled nodes.

Let \(V_U\) denote the set of unlabeled nodes. Nodes with a maximum class probability exceeding a confidence threshold \(\tau\) are selected as pseudo-labeled:
\begin{equation}
V_{\text{pseudo}} = \left\{ v_i \in V_U \;\middle|\; \max(\hat{y}_i) > \tau \right\}
\end{equation}

The extended labeled set is formed as:
\begin{equation}
V_L' = V_L \cup V_{\text{pseudo}}, \quad y_L' = y_L \cup \left\{ \arg\max \hat{y}_i \mid v_i \in V_{\text{pseudo}} \right\}
\end{equation}

The model is retrained by minimizing a composite loss function that balances supervised classification, sparsity control, and temporal smoothness:
\begin{equation}
L_{\text{total}} = L_{\text{sup}} + \lambda_1 L_{\text{sparsity}} + \lambda_2 L_{\text{temporal}}
\end{equation}
Here, \(L_{\text{sup}}\) is the cross-entropy loss over \(V_L'\). The sparsity regularization \(L_{\text{sparsity}}\) penalizes large edge weights to avoid overfitting and promote compact graph structure:
\begin{equation}
L_{\text{sparsity}} = \sum_{(i,j) \in E} \| w_{ij} \|_1
\end{equation}

The temporal smoothness loss \(L_{\text{temporal}}\) ensures consistent relationships between temporally adjacent segments:
\begin{equation}
L_{\text{temporal}} = \sum_{(i,j),\,(i,j+1) \in E} (w_{ij} - w_{i,j+1})^2
\end{equation}
This iterative self-training strategy enables the model to gradually expand supervision from reliable predictions while maintaining structural sparsity and temporal coherence. The complete processing pipeline is summarized in Algorithm~\ref{alg:ss_asd}.

\begin{algorithm}[tb]
\caption{Semi-Supervised Abnormal Stop Detection (ASD)}
\label{alg:ss_asd}
\begin{algorithmic}[1]

\Require Sparse GPS trajectory $T = \{p_1, p_2, \dots, p_n\}$

\State Compute adaptive thresholds \hfill \textit{(Eq.~8)}
\State Segment $T$ where either $D_{ij} > \lambda_d$ or $\Delta t_{ij} > \lambda_t$

\For{each segment $r_i$}
    \State Compute indicators: $\text{TIS}_{ij}$, $\text{MSD}_i$, $\text{TTA@k}_i$ \hfill \textit{(Eq.~10--12)}
    \State Estimate segment confidence $C_i$ \hfill \textit{(Eq.~13)}
    
    \If{$C_i < \tau_c$}
    \State Apply $\lambda$-smoothing: $\mathbf{x}_i' \leftarrow$ LTIGA \hfill \textit{(Eq.~14,15)}
    \State Inverse normalize: $\mathbf{x}_i \leftarrow \mathbf{x}_i' \cdot \boldsymbol{\sigma}_k + \boldsymbol{\mu}_k$\hfill \textit{(Eq.~16)} 
    \State Apply confidence reweighting: $\mathbf{x}_i^{\text{w}} \leftarrow C_i \cdot \mathbf{x}_i$ \hfill \textit{(Eq.~17)}
    \Else
    \State Use raw indicators: $\mathbf{x}_i^{\text{w}} \leftarrow \mathbf{x}_i$
    \EndIf
\EndFor

\State Construct graph $G = (V, E)$ from $\{\mathbf{x}_i^{\text{w}}\}$
\State Perform label propagation using $Q(F)$ \hfill \textit{(Eq.~20)}

\For{each node $i$}
    \If{$\max(\hat{y}_i) \geq \tau_1$ or $\leq \tau_0$}
        \State Assign pseudo-label $\hat{y}_i$ \hfill \textit{(Eq.~23)}
    \EndIf
\EndFor

\State Train GCN on pseudo-labeled nodes using $L_{\text{GCN}}$ \hfill \textit{(Eq.~25)}
\State Update predictions via self-training: $\hat{y}_i \leftarrow \hat{y}_i^{(t+1)}$

\Ensure Final abnormal stop predictions $\hat{y}_i$

\end{algorithmic}
\end{algorithm}

\section{Data Analysis and Experimental Results} \label{sec:dataset_analysis}

\subsection{\textbf{Data}}
The GPS data were collected along a coach route between Liuliqiao Station in Beijing and Zhangjiakou Station in Hebei Province for our lab \cite{deng2024unsupervised}. Table~\ref{tab:dataset_summary} summarizes the attributes of each trajectory point.

\begin{table}[tb]
\centering
\caption{Summary of Dataset Attributes for Vehicle Trajectory Analysis}
\resizebox{\columnwidth}{!}{
\begin{tabular}{|l|l|l|}
\hline
\textbf{Attribute Name} & \textbf{Category} & \textbf{Data Type} \\ \hline
\multicolumn{3}{|c|}{\textbf{Spatial Information}} \\ \hline
Longitude (lng) & Spatial & Continuous (Decimal Degrees) \\ \hline
Latitude (lat) & Spatial & Continuous (Decimal Degrees) \\ \hline
\multicolumn{3}{|c|}{\textbf{Vehicle Movement Data}} \\ \hline
Stay Time (s) & Temporal & Continuous (Seconds) \\ \hline
Speed (v) & Movement & Continuous (km/h) \\ \hline
Distance from Previous Point & Movement & Continuous (Meters) \\ \hline
Distance from Starting Point (d) & Movement & Continuous (Meters) \\ \hline
\multicolumn{3}{|c|}{\textbf{Temporal Data}} \\ \hline
Date (date) & Temporal & Date/Time (Timestamp) \\ \hline
Time (t) & Temporal & Continuous (Seconds) \\ \hline
\end{tabular}
}
\label{tab:dataset_summary}
\end{table}

\subsection{\textbf{Discriminative Power of Domain-Specific Indicators}}

Prior to any refinement, we evaluate the discriminative power of TIS, MSD, and TTA@k on the raw GPS dataset to assess their standalone effectiveness in identifying abnormal segments.

As shown in Table~\ref{tab:indicator_performance}, while the AUC values are modest (e.g., TIS: 0.26, MSD: 0.49, TTA@k: 0.29), the AP scores are substantially higher across all indicators (e.g., AP$_{\text{TIS}}$: 0.95, AP$_{\text{MSD}}$: 0.96, AP$_{\text{TTA@k}}$: 0.95). This reflects a common phenomenon in imbalanced classification: AUC, which measures class separability, tends to underperform when positive cases are rare—as in our dataset, which contains only 10 ground-truth abnormal labels—while AP, which emphasizes ranking quality, better captures the model’s ability to prioritize true positives.

The high AP scores suggest that these indicators are behaviorally meaningful and effective at ranking true abnormal stops higher in the output—despite the sparsity and noise inherent in GPS trajectories. This makes them particularly well-suited to guide downstream learning tasks—such as LTIGA. 

\subsection{\textbf{Impact of LTIGA on Indicator Discrimination}}

To assess the contribution of LTIGA under conditions of label scarcity, indicator performance is compared before and after its integration. Table~\ref{tab:indicator_performance} reports AUC and AP scores for indicators. TIS and TTA@k show the most substantial gains in AUC, while MSD shows marginal
improvement. These improvements are particularly meaningful given the extreme imbalance of only 10 ground-truth abnormal stops, underscoring LTIGA’s capacity to enhance indicator expressiveness without overfitting.

\begin{table}[H]
\centering
\caption{Indicator Performance Before and After LTIGA Refinement}
\label{tab:indicator_performance}
\begin{tabular}{lcccc}
\toprule
\textbf{Indicator} & \textbf{AUC (Before)} & \textbf{AUC (After)} & \textbf{AP (Before)} & \textbf{AP (After)} \\
\midrule
TIS     & 0.2673 & \textbf{0.4392} & 0.950 & 0.950 \\
MSD     & 0.4851 & \textbf{0.4916} & 0.960 & 0.940 \\
TTA@k   & 0.2986 & \textbf{0.4647} & 0.940 & 0.940 \\
\bottomrule
\end{tabular}
\end{table}

\subsection{\textbf{Enhancing Supervision Coverage via Label Propagation}}

To increase effective supervision while preserving label quality, label propagation was applied over the constructed graph with LTIGA-refined features. The process selectively expanded labels using an RBF kernel affinity matrix while enforcing high-confidence thresholds (\(\geq 0.995\) for normal, \(\leq 0.005\) for abnormal) and capping the number of abnormal pseudo-labels to maintain class purity.

Prior to propagation, 523 nodes were labeled, comprising 10 abnormal and 513 normal cases. After propagation, the labeled set expanded to 712 nodes, with 12 abnormal and 700 normal labels retained, as summarized in Table~\ref{tab:label_propagation_summary}. This controlled expansion preserved the rarity of abnormal cases and avoided oversaturation, while still improving labeled sample availability for GCN training. Importantly, 4884 out of 5596 nodes remained unlabeled, maintaining the integrity of the semi-supervised learning setup.

\begin{table}[H]
\centering
\caption{Label Distribution After Label Propagation}
\label{tab:label_propagation_summary}
\begin{tabular}{l c p{3.0cm}}
\toprule
\textbf{Class} & \textbf{After Propagation} & \textbf{Justification} \\
\midrule
Abnormal (Class 0) & 10--12 & Maintain rarity for abnormal stop detection\\
Normal (Class 1) & 513--700 & Expanded for stronger supervision while avoiding imbalance \\
Unlabeled & 5073--5200 & Preserve majority for SSL \\
\bottomrule
\end{tabular}
\end{table}

\subsection{Evaluation of ASD with Graph-Based SSL}

We evaluate ASD using two metrics: Area Under the ROC Curve (AUC) and Average Precision (AP), both commonly used in imbalanced classification tasks. AUC quantifies separability between classes, while AP emphasizes the model’s ranking performance for rare positive cases.

\begin{table}[H]
\centering
\caption{Performance Comparison with Existing Methods}
\label{tab:sota_comparison}
\begin{tabular}{lcc}
\toprule
\textbf{Method} & \textbf{AUC} & \textbf{AP} \\
\midrule
WST & 0.7143 & 0.5317 \\
WSA & 0.5238 & 0.3694 \\
UIS & 0.5312 & 0.2678 \\
Unsupervised ASD~\cite{deng2024unsupervised} & 0.7619 & 0.5556 \\
\textbf{Ours (GCN)} & \textbf{0.8542} & \textbf{0.8661} \\
\bottomrule
\end{tabular}
\end{table}

The GCN model, trained on label-propagated data, achieved a strong AUC of 0.8542 and AP of 0.8661—significantly outperforming prior methods (Table~\ref{tab:sota_comparison}). Although predictions are made at the node level, final abnormality decisions are aggregated at the segment level to align with ASD’s practical objective. Segment-level aggregation ensures that even partially abnormal segments are flagged appropriately.

A self-training phase was subsequently applied to improve generalization by augmenting supervision with high-confidence pseudo-labels. The results before and after refinement are summarized in Table~\ref{tab:st_results}. Self-training improves both discrimination and coverage, with AUC rising to 0.8819 and AP to 0.8842.

\begin{table}[H]
\centering
\caption{GCN Performance Before and After Self-Training}
\label{tab:st_results}
\begin{tabular}{lcc}
\toprule
\textbf{Metric} & \textbf{Before ST} & \textbf{After ST} \\
\midrule
AUC & 0.8542 & \textbf{0.8819} \\
Average Precision (AP) & 0.8661 & \textbf{0.8842} \\
Abnormal Nodes Detected & 43 & 84 \\
Abnormal Segments Identified & 28 & 50 \\
\bottomrule
\end{tabular}
\end{table}

\subsection{Spatial Validation of Predicted Abnormal Stops}

\begin{figure}[H]
\centering
\begin{subfigure}[b]{0.31\columnwidth}
    \includegraphics[width=\linewidth]{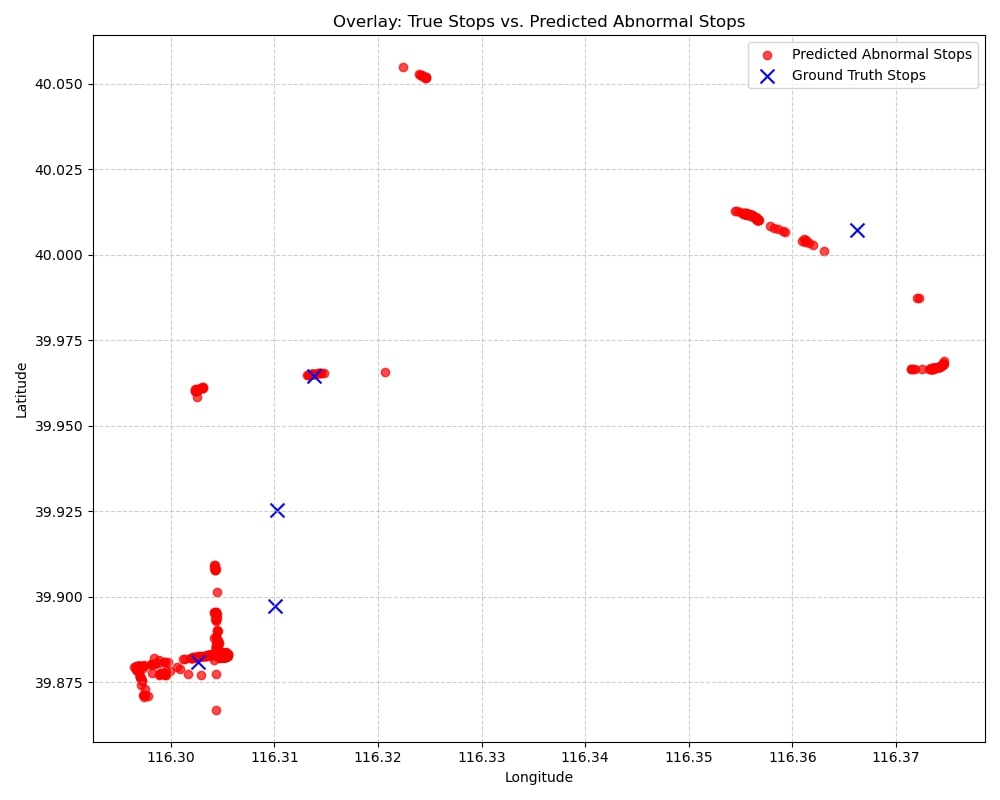}
    \caption{\(k=5\), Trial 1}
    \label{fig:overlay_k5}
\end{subfigure}
\hfill
\begin{subfigure}[b]{0.31\columnwidth}
    \includegraphics[width=\linewidth]{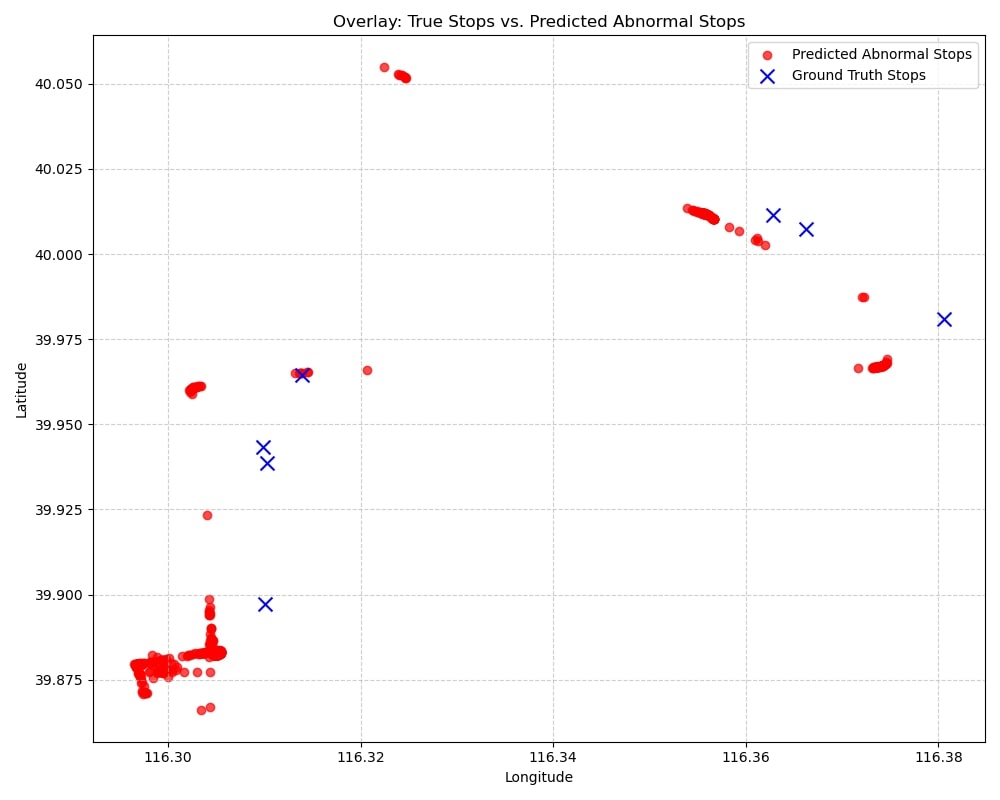}
    \caption{\(k=7\), Trial 2}
    \label{fig:overlay_k7}
\end{subfigure}
\hfill
\begin{subfigure}[b]{0.31\columnwidth}
    \includegraphics[width=\linewidth]{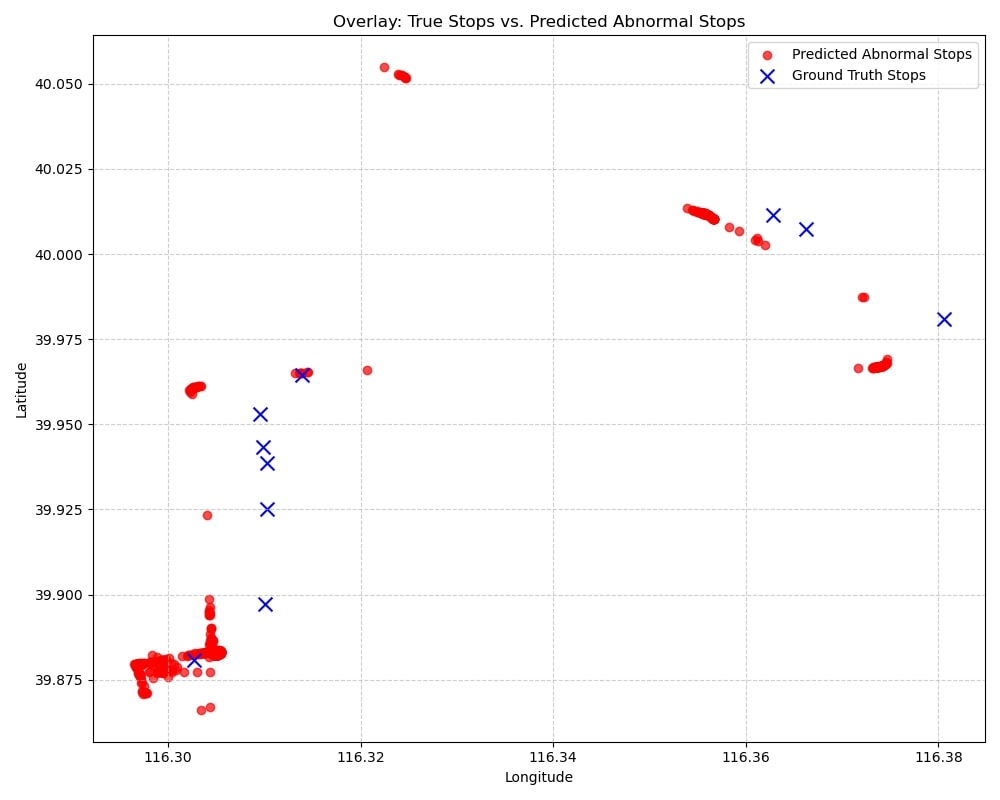}
    \caption{\(k=10\), Trial 2}
    \label{fig:overlay_k10}
\end{subfigure}
\caption{Spatial overlays of predicted vs. ground-truth abnormal stops under three label budgets. Prediction improves with more supervision—from failure (a) to moderate (b) to high-quality detection (c).}
\label{fig:fig3_overlays}
\vspace{-1mm}
\end{figure}

To assess the spatial accuracy and label efficiency of the proposed framework, a qualitative and quantitative validation was performed on predicted abnormal stops under varying label budgets \( k \in \{5, 7, 10\} \). For each \( k \), five independent trials were conducted with randomly sampled labeled stops. Figure~\ref{fig:fig3_overlays} presents representative overlays for one selected trial at each \( k \), chosen to illustrate progressively improved detection performance. The \(k=5\) case highlights a failure scenario, where prediction coverage is limited—reflecting the model's uncertainty under extremely sparse supervision. In contrast, the \(k=7\) and \(k=10\) cases exhibit increasingly consistent alignment between predicted anomalies and ground-truth stop locations, demonstrating the framework’s ability to generalize abnormal stop patterns with modest increases in labeled data. 

To complement the visual overlays, Table~\ref{tab:label_variation_summary} summarizes the detection performance, number of abnormal segments recovered, and spatial error statistics for the selected trials.

\begin{table}[ht]
\centering
\caption{Representative trials for abnormal stop detection with varying label counts ($k$)}
\label{tab:label_variation_summary}
\resizebox{\columnwidth}{!}{
\begin{tabular}{ccccccc}
\toprule
\textbf{$k$} & \textbf{Trial} & \textbf{AUC} & \textbf{AP} & \textbf{Segments} & \textbf{Mean Dist.\ (m)} & \textbf{Med.\ Dist.\ (m)} \\
\midrule
5  & 1 & 0.8827 & 0.9054 & 35 & 618.64 & 514.78 \\
7  & 2 & 0.8571 & 0.8100 & 37 & 892.46 & 537.11 \\
10 & 2 & 0.8571 & 0.8100 & 37 & 783.75 & 554.61 \\
\bottomrule
\end{tabular}
}
\vspace{-1mm}
\end{table}

While the visual overlays in Fig.~\ref{fig:fig3_overlays} may appear similar at first glance—particularly in dense urban regions where abnormal stops cluster spatially—the underlying predictions differ across trials. These subtle differences, though visually occluded, are quantitatively reflected in Table~\ref{tab:label_variation_summary}, which highlights the impact of label budget on model behavior. For a more comprehensive perspective, the supplementary material includes all five overlay maps for each label budget \(k\). It also provides full performance metrics across trials, further confirming the robustness and consistency of our framework under varying label configurations.

In addition to the qualitative overlay, a quantitative analysis was conducted to measure the spatial proximity between predicted abnormal segments and ground-truth stops. Specifically, for each ground-truth stop, the minimum distance to the nearest predicted abnormal segment was computed. As summarized in Table~\ref{tab:distance_comparison}, the distances demonstrate that a considerable number of predictions occur within a reasonable spatial neighborhood of actual abnormal events, despite the model being trained without access to coordinate-level supervision. 

\begin{table}[H]
\scriptsize 
\renewcommand{\arraystretch}{1.1}
\caption{Distances Between Ground Truth and Predicted Stops}
\label{tab:distance_comparison}
\centering
\begin{tabular}{|c|c|c|c|}
\hline
\textbf{Stop} & \textbf{Ground Truth (Lon, Lat)} & \textbf{Predicted (Lon, Lat)} & \textbf{Distance (km)} \\
\hline
1 & (116.3027, 39.8809) & (116.3020, 39.8821) & 0.14 \\
2 & (116.3101, 39.8973) & (116.3044, 39.8957) & 0.52 \\
3 & (116.3103, 39.9253) & (116.3042, 39.9093) & 1.85 \\
4 & (116.3103, 39.9387) & (116.3025, 39.9584) & 2.29 \\
5 & (116.3098, 39.9434) & (116.3025, 39.9584) & 1.78 \\
6 & (116.3096, 39.9532) & (116.3025, 39.9584) & 0.84 \\
7 & (116.3139, 39.9644) & (116.3136, 39.9650) & 0.07 \\
8 & (116.3806, 39.9810) & (116.3746, 39.9691) & 1.42 \\
9 & (116.3662, 40.0072) & (116.3613, 40.0042) & 0.54 \\
10 & (116.3628, 40.0115) & (116.3565, 40.0110) & 0.54 \\
\hline
\textbf{Mean} & -- & -- & \textbf{1.10} \\
\textbf{Median} & -- & -- & \textbf{0.69} \\
\hline
\end{tabular}
\end{table}

The mean distance between matched points was 1.10 kilometers, while the median distance was 0.69 kilometers, indicating that the nearest predicted abnormalities are geographically close to real abnormal stop locations. Note that a single predicted abnormal stop may correspond to multiple ground-truth stops if it is the nearest prediction for more than one true stop location. These distance measurements provide a concrete assessment of the spatial closeness between predicted abnormalities and true stop locations.

\begin{figure}[H]
  \centering
  \includegraphics[width=\linewidth]{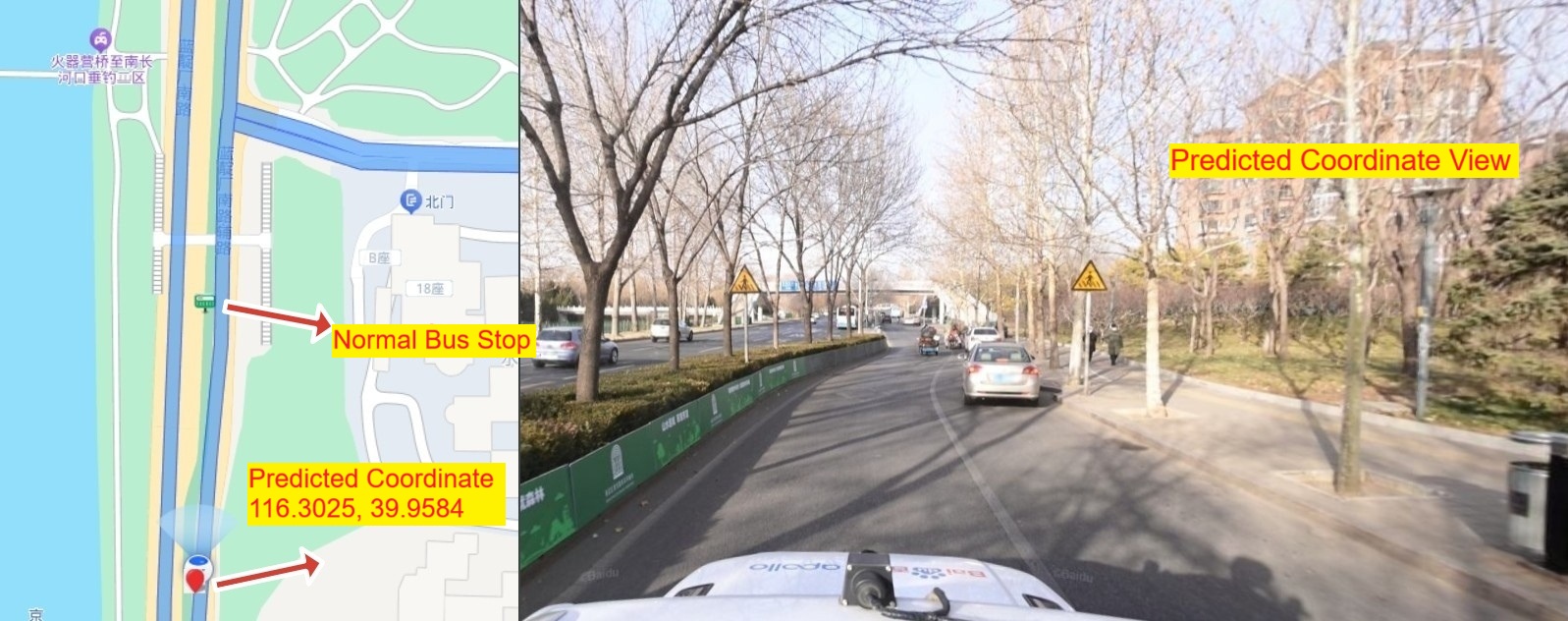}
  \caption{Map and street view of predicted stop at (116.3025, 39.9584).}
  \label{fig:streetview_comparison_1}
\end{figure}

\begin{figure}[H]
  \centering
  \includegraphics[width=\linewidth]{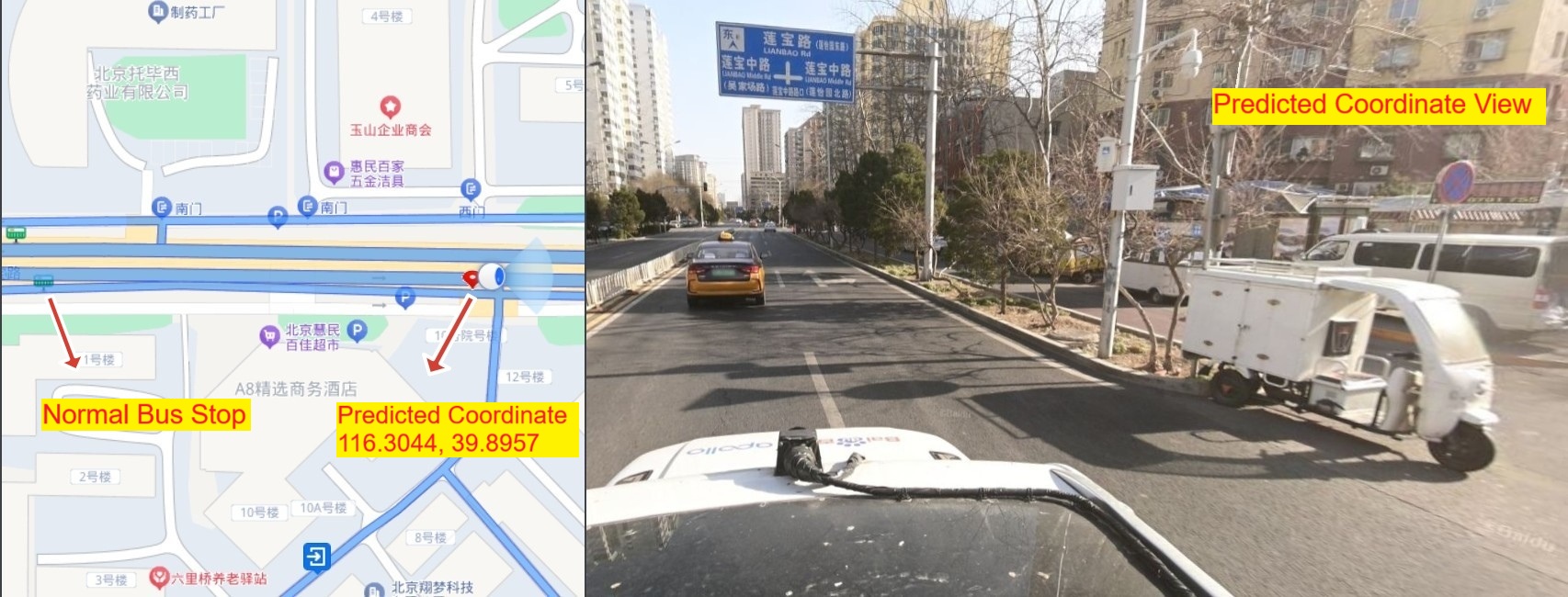}
  \caption{Map and street view of predicted stop at (116.3044, 39.8957).}
  \label{fig:streetview_comparison_2}
\end{figure}

To provide contextual justification for the predicted abnormalities listed in Table~\ref{tab:distance_comparison}, we visualize two representative cases using corresponding map and street view imagery. Fig.~\ref{fig:streetview_comparison_1} and Fig.~\ref{fig:streetview_comparison_2} depict the predicted coordinates (116.3025, 39.9584) and (116.3044, 39.8957), respectively, alongside their nearby annotated normal stops. In both cases, the predicted points appear in informal roadside areas without marked infrastructure, illustrating the model’s ability to capture subtle deviations from expected stop locations.

\graphicspath{{figures/}}

\subsection{Impact of Label Quantity on Performance}

To evaluate the framework's robustness under sparse supervision, we conduct a sensitivity analysis by varying the number of labeled abnormal GPS stops from 5 to 10. As shown in Table~\ref{tab:label_sensitivity_summary}, the model maintains stable and high-quality performance across different label budgets. Notably, the improvements in AUC and AP begin to plateau beyond 7 labels, indicating diminishing returns with additional supervision. Moreover, the steady increase in detected abnormal segments and the reduction in median localization error reflect improved spatial precision with more labels. These results confirm that the framework is label-efficient, maintaining stable classification performance and precise spatial detection under limited annotation.

\begin{table}[H]
\centering
\caption{Label Sensitivity Analysis Across Varying Numbers of GPS Stop Labels}
\label{tab:label_sensitivity_summary}
\resizebox{\columnwidth}{!}{
\begin{tabular}{ccccccc}
\toprule
\textbf{\# Labels} & \textbf{AUC (mean$\pm$std)} & \textbf{AP (mean$\pm$std)} & \textbf{Abn.\ Nodes} & \textbf{Abn.\ Segments} & \textbf{Mean Dist.\ (m)} & \textbf{Median Dist.\ (m)} \\
\midrule
5  & 0.1765 ± 0.3948 & 0.1811 ± 0.4049 & 17.4  & 7.0  & 903.63 & 628.77 \\
7  & 0.6743 ± 0.3782 & 0.6849 ± 0.3844 & 68.4  & 22.0 & 987.47 & 802.43 \\
10 & 0.8507 ± 0.0356 & 0.8674 ± 0.0423 & 70.6  & 23.0 & 1170.45 & 1161.96 \\
\bottomrule
\end{tabular}
}
\vspace{-2mm}
\end{table}

\subsection{Ablation Study}

\subsubsection{Ablation on SAS: Sparsity-Aware vs. Fixed-Length Segmentation}

To assess the contribution of our proposed Sparsity-Aware Segmentation (SAS), we compare the full framework against a variant that replaces SAS with fixed-length segmentation, where trajectories are uniformly partitioned into 2 km intervals. As shown in Table~\ref{tab:ablation_sas}, removing SAS significantly degrades detection performance, increases the number of abnormal predictions, and reduces graph connectivity—highlighting the importance of adaptive segmentation.

\begin{table}[H]
\centering
\caption{Impact of SAS vs. Fixed-Length Segmentation}
\label{tab:ablation_sas}
\resizebox{\columnwidth}{!}{
\begin{tabular}{lcccccc}
\toprule
\textbf{Variant} & \textbf{AUC$_\text{GCN}$} & \textbf{AP$_\text{GCN}$} & \textbf{Abn.\ Nodes} & \textbf{Segments} & \textbf{Med.\ Dist.\ (km)} & \textbf{Graph Edges} \\
\midrule
SAS + LTIGA   & 0.854 & 0.866 & 43  & 22  & 0.69 & 4651 \\
Fixed + LTIGA & 0.799 & 0.786 & 110 & 60  & 0.76 & 1990 \\
\bottomrule
\end{tabular}
}
\vspace{-1mm}
\end{table}

\begin{figure}[H]
  \centering
  \begin{subfigure}[b]{0.48\columnwidth}
    \includegraphics[width=\linewidth]{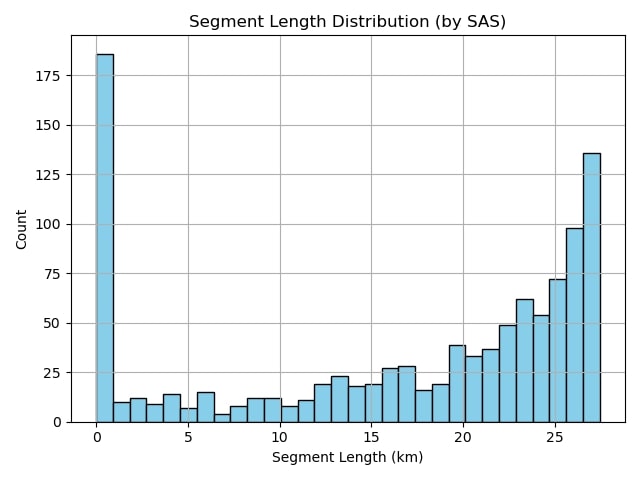}
    \caption{SAS: Segment length}
  \end{subfigure}
  \hfill
  \begin{subfigure}[b]{0.48\columnwidth}
    \includegraphics[width=\linewidth]{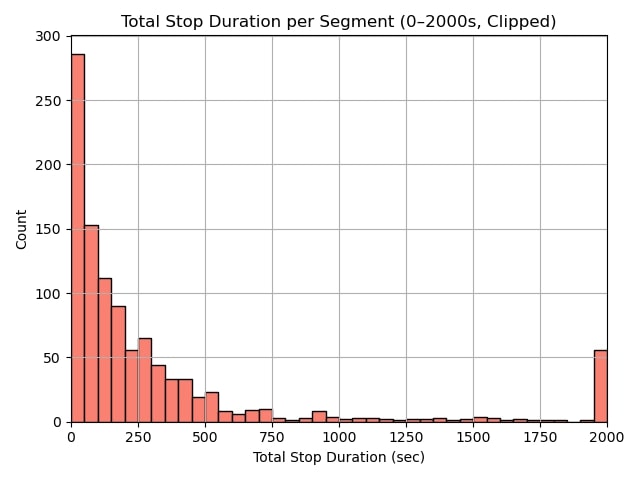}
    \caption{SAS: Stop duration}
  \end{subfigure}
  \begin{subfigure}[b]{0.48\columnwidth}
    \includegraphics[width=\linewidth]{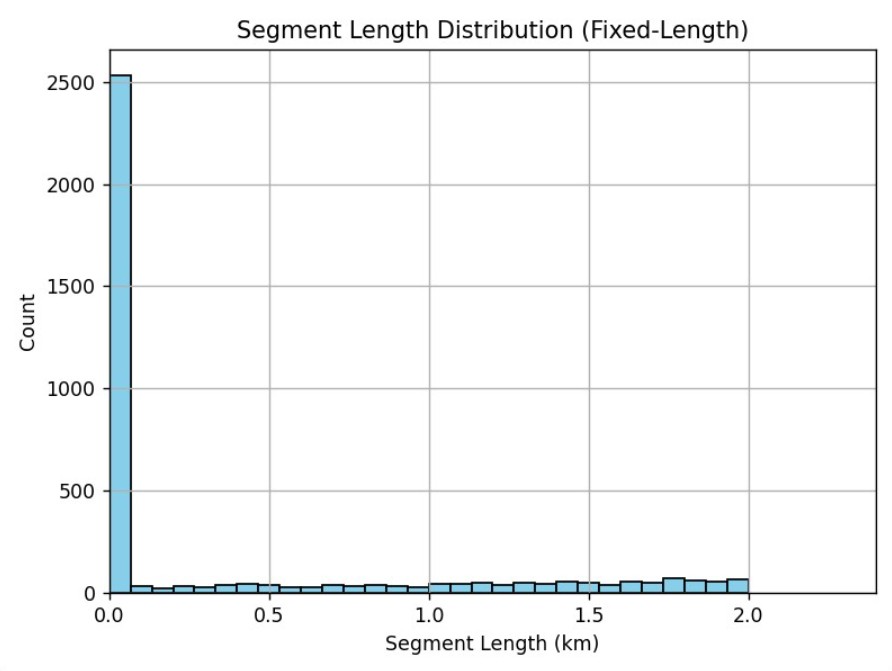}
    \caption{Fixed: Segment length}
  \end{subfigure}
  \hfill
  \begin{subfigure}[b]{0.48\columnwidth}
    \includegraphics[width=\linewidth]{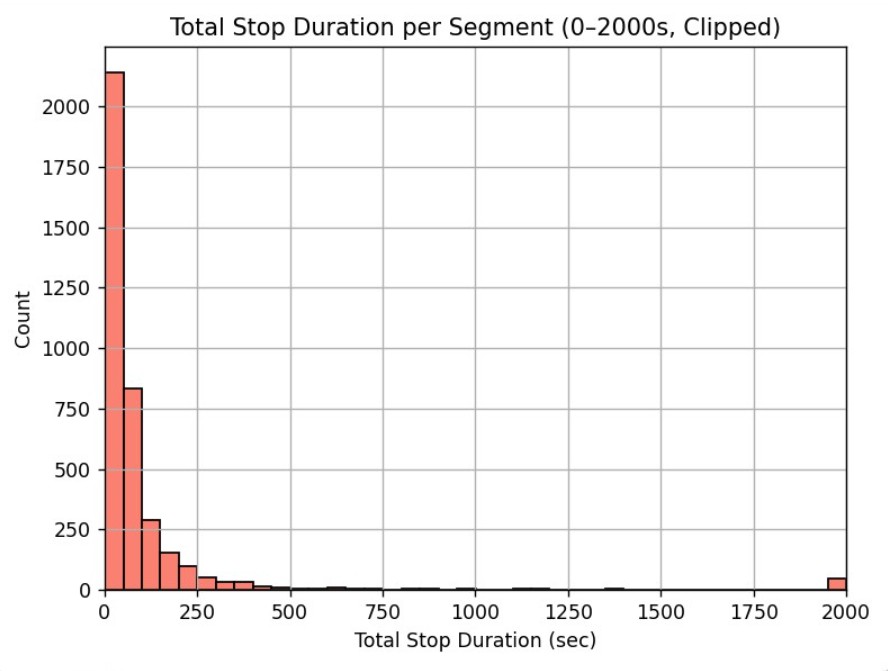}
    \caption{Fixed: Stop duration}
  \end{subfigure}
  \caption{Comparison of segmentation distributions under SAS and fixed-length segmentation}
  \label{fig:sas_vs_fixed}
\end{figure}

Figure~\ref{fig:sas_vs_fixed} further illustrates the structural differences in segment distributions. The top row corresponds to SAS. In Figure~\ref{fig:sas_vs_fixed}a, segment lengths range from under 1 km to over 25 km, with peaks at both extremes. This variability reflects SAS's ability to adapt to heterogeneous GPS sampling densities—producing shorter segments in dense urban areas and longer ones in sparse regions—without relying on manual threshold tuning. 

Figure~\ref{fig:sas_vs_fixed}b shows total stop duration per segment under SAS. While most segments correspond to routine stops, a noticeable long-tail remains even after clipping at 2000 seconds for readability. These extended durations preserve critical behavioral signals that are essential for identifying abnormal events.

In contrast, the bottom row presents results from fixed-length segmentation. This method uniformly slices trajectories, producing over 2500 short and homogeneous segments concentrated around 0.2 km in length (Figure~\ref{fig:sas_vs_fixed}c). Correspondingly, the total stop durations per segment (Figure~\ref{fig:sas_vs_fixed}d) are highly compressed, exhibiting limited behavioral variability. Although all histograms share consistent axis limits to ensure fair comparison, the distributional differences are striking—underscoring the advantage of SAS in preserving real-world travel dynamics and heterogeneity.

\subsubsection{Ablation on LTIGA Smoothing}

To evaluate the contribution of LTIGA, we compare the proposed model with a variant in which the LTIGA smoothing step is removed. As presented in Table~\ref{tab:ablation_ltiga}, disabling LTIGA leads to a noticeable drop in detection performance and a sharp increase in falsely predicted abnormal nodes. While the median spatial distance to ground-truth stops appears marginally reduced, this is largely due to over-prediction rather than improved localization.

\begin{table}[ht]
\centering
\caption{Impact of LTIGA smoothing on model performance and prediction quality}
\label{tab:ablation_ltiga}
\resizebox{\columnwidth}{!}{
\begin{tabular}{lccccc}
\toprule
\textbf{Variant} & \textbf{AUC$_\text{GCN}$} & \textbf{AP$_\text{GCN}$} & \textbf{\# Abn.\ Nodes} & \textbf{\# Segm.} & \textbf{Med.\ Dist.\ (km)} \\
\midrule
With LTIGA      & 0.854 & 0.866 & 43  & 22 & 0.69 \\
w/o LTIGA       & 0.771 & 0.813 & 124 & 41 & 0.56 \\
\bottomrule
\end{tabular}
}
\vspace{-3mm}
\end{table}
\subsection{Ablation Study on Semantic Restoration (Eq. 16)}

To evaluate the contribution of the semantic restoration step (Eq. 16) in our LTIGA module, we perform an ablation study comparing two configurations: (i) LTIGA without semantic restoration, where the smoothed indicators remain in normalized (z-score) form; and (ii) LTIGA with semantic restoration (\textbf{Ours}), where smoothed values are rescaled back to their original distributions using $\mathbf{x}_i^{\text{smooth}} = \tilde{\mathbf{x}}_i' \cdot \boldsymbol{\sigma}_k + \boldsymbol{\mu}_k  $. This rescaling ensures that the denoised indicators retain their real-world semantic meaning (e.g., stop duration in seconds, movement density in meters), which is critical for downstream graph embedding and pseudo-label propagation. Without this step, the GCN operates on context-free z-scores, which can obscure true behavioral distinctions—particularly under uneven sampling or sparse GPS data. Table~\ref{tab:ablation_rescaling} presents a comparison between the two variants.

\begin{table}[ht]
\centering
\caption{Ablation of Semantic Restoration Step (Eq. 16)}
\label{tab:ablation_rescaling}
\resizebox{\columnwidth}{!}{
\begin{tabular}{lccccc}
\toprule
\textbf{Variant} & \textbf{AUC} & \textbf{AP} & \textbf{Abn.\ Nodes} & \textbf{Match Dist.\ (km)} & \textbf{Segments} \\
\midrule
LTIGA (w/o Rescaling) & 0.6597 & 0.7342 & 56 & 0.50 & 80 \\
LTIGA (with Rescaling, Ours) & \textbf{0.8542} & \textbf{0.8661} & \textbf{43} & \textbf{1.00} & \textbf{22} \\
\bottomrule
\end{tabular}
}
\vspace{-1mm}
\end{table}

Although the non-rescaled variant detects more abnormal nodes and segments with slightly lower matching distance, this reflects over-detection due to loss of semantic context. In contrast, our method detects fewer but more behaviorally meaningful anomalies, yielding substantially higher AUC and AP. This confirms that semantic restoration (Eq. 16) improves both precision and reliability of abnormal stop detection.

\subsubsection{GCN Baseline Analysis}

To isolate the effect of our complete pipeline, we evaluate a simplified GCN baseline—excluding both SAS and LTIGA. As shown in Table~\ref{tab:ablation_gcnonly}, performance declines substantially across all metrics, with lower AUC/AP, inflated abnormal node count, over-fragmented segments and over 2500 skipped nodes due to unreliable features. 

\begin{table}[ht]
\centering
\caption{Comparison of our model vs.\ GCN baseline across detection performance and graph connectivity metrics}
\label{tab:ablation_gcnonly}
\resizebox{\columnwidth}{!}{
\begin{tabular}{lccccccc}
\toprule
\textbf{Variant} & \textbf{AUC} & \textbf{AP} & \textbf{Abn.\ Nodes} & \textbf{Segments} & \textbf{Med.\ Dist.\ (km)} & \textbf{Skipped Nodes} & \textbf{Edges} \\
\midrule
GCN (SAS + LTIGA) & 0.854 & 0.866 & 43  & 22  & 0.69 & 212 & 4651 \\
GCN-only (Raw+ Fixed)     & 0.653 & 0.689 & 115 & 60  & 0.58 & 2503 & 1990 \\
\bottomrule
\end{tabular}
}
\vspace{-2mm}
\end{table}

\section{Limitations and Future Work}

The proposed framework demonstrates robust performance under sparse GPS conditions and extreme label scarcity. The use of domain-specific, handcrafted indicators provides interpretability and stability in low-label scenarios, making the model suitable for deployment in real-world transportation systems. Nevertheless, future work could explore the integration of learned trajectory representations, such as pretrained spatiotemporal transformers, to capture richer behavioral patterns. Additionally, segment-level modeling effectively reduces noise but assumes behavioral consistency within each segment. In cases where intra-segment variation is high, finer-grained modeling may offer improved localization of subtle anomalies. Furthermore, while the graph construction and label propagation components perform well with hyperparameters, adaptive strategies based on context or uncertainty could further enhance generalization across diverse routes. Future research may also extend the framework to multi-route environments and explore real-time capabilities using efficient, streaming-compatible graph architectures.

\section{Conclusion}
This paper presents a semi-supervised graph-based framework for detecting abnormal stops in sparse GPS trajectories, specifically addressing the challenges of low sampling frequency and extreme label scarcity in long-distance coach transportation. The proposed pipeline combines sparsity-aware segmentation (SAS), indicator refinement through LTIGA, confidence-weighted graph construction, label propagation, and iterative self-training. These components collectively enable accurate, interpretable anomaly detection with minimal supervision. Extensive experiments show that the method consistently outperforms existing baselines in both AUC and AP, while effectively localizing true anomalies under sparse data conditions. The framework's modular design, label efficiency, and robustness to noise make it well-suited for deployment in real-world transportation systems with limited sensing infrastructure and annotation budgets.

\begingroup
\small 
\bibliographystyle{IEEEtran}
\bibliography{ref}

\begin{thebibliography}{10}
\providecommand{\url}[1]{#1}
\csname url@samestyle\endcsname
\providecommand{\newblock}{\relax}
\providecommand{\bibinfo}[2]{#2}
\providecommand{\BIBentrySTDinterwordspacing}{\spaceskip=0pt\relax}
\providecommand{\BIBentryALTinterwordstretchfactor}{4}
\providecommand{\BIBentryALTinterwordspacing}{\spaceskip=\fontdimen2\font plus
\BIBentryALTinterwordstretchfactor\fontdimen3\font minus \fontdimen4\font\relax}
\providecommand{\BIBforeignlanguage}[2]{{%
\expandafter\ifx\csname l@#1\endcsname\relax
\typeout{** WARNING: IEEEtran.bst: No hyphenation pattern has been}%
\typeout{** loaded for the language `#1'. Using the pattern for}%
\typeout{** the default language instead.}%
\else
\language=\csname l@#1\endcsname
\fi
#2}}
\providecommand{\BIBdecl}{\relax}
\BIBdecl

\bibitem{oh2024enhancing}
S.~Oh and J.~Cho, ``Enhancing regional connectivity: A multimodal accessibility assessment for public service obligation routes,'' \emph{Transportation Research Part D: Transport and Environment}, vol. 134, p. 104333, 2024.

\bibitem{fan2019research}
Z.~Fan, C.~Liu, D.~Cai, and S.~Yue, ``Research on black spot identification of safety in urban traffic accidents based on machine learning method,'' \emph{Safety science}, vol. 118, pp. 607--616, 2019.

\bibitem{zhang2014understanding}
D.~Zhang, L.~Sun, B.~Li, C.~Chen, G.~Pan, S.~Li, and Z.~Wu, ``Understanding taxi service strategies from taxi gps traces,'' \emph{IEEE Transactions on Intelligent Transportation Systems}, vol.~16, no.~1, pp. 123--135, 2014.

\bibitem{de2024banning}
A.~de~Bortoli and A.~F{\'e}raille, ``Banning short-haul flights and investing in high-speed railways for a sustainable future?'' \emph{Transportation Research Part D: Transport and Environment}, vol. 128, p. 103987, 2024.

\bibitem{nature2024longdistance}
L.~Jin, X.~Zhao, and Y.~Zhou, ``The importance of long-distance travel in decarbonizing passenger transport,'' \emph{Nature Energy}, vol.~9, pp. 421--430, 2024.

\bibitem{chen2013iboat}
C.~Chen, D.~Zhang, P.~S. Castro, N.~Li, L.~Sun, S.~Li, and Z.~Wang, ``iboat: Isolation-based online anomalous trajectory detection,'' \emph{IEEE Transactions on Intelligent Transportation Systems}, vol.~14, no.~2, pp. 806--818, 2013.

\bibitem{stokenberga2025connecting}
A.~Stokenberga, E.~Sa{\"\i}sset, T.~Kerzhner, and X.~Espinet~Alegre, ``Connecting through public transport: accessibility to health and education in major african cities,'' \emph{Area Development and Policy}, pp. 1--18, 2025.

\bibitem{van2020preferences}
V.~Van~Acker, R.~Kessels, D.~P. Cuervo, S.~Lannoo, and F.~Witlox, ``Preferences for long-distance coach transport: Evidence from a discrete choice experiment,'' \emph{Transportation Research Part A: Policy and Practice}, vol. 132, pp. 759--779, 2020.

\bibitem{deng2024unsupervised}
J.~Deng, J.~Pang, J.~Xu, and H.~Yu, ``Unsupervised abnormal stop detection for long distance coaches with low-frequency gps,'' \emph{arXiv preprint arXiv:2411.04422}, 2024.

\bibitem{pang2017discovering}
J.~Pang, J.~Huang, X.~Yang, Z.~Wang, H.~Yu, Q.~Huang, and B.~Yin, ``Discovering fine-grained spatial pattern from taxi trips: Where point process meets matrix decomposition and factorization,'' \emph{IEEE Transactions on Intelligent Transportation Systems}, vol.~19, no.~10, pp. 3208--3219, 2017.

\bibitem{xu2024perimeter}
Y.~Xu, D.~Li, and Y.~Xi, ``Perimeter traffic flow control for a multi-region large-scale traffic network with markov decision process,'' \emph{IEEE Transactions on Intelligent Transportation Systems}, vol.~25, no.~6, pp. 4809--4821, 2024.

\bibitem{ceccato2022crime}
V.~Ceccato, N.~Gaudelet, and G.~Graf, ``Crime and safety in transit environments: a systematic review of the english and the french literature, 1970--2020,'' \emph{Public Transport}, vol.~14, no.~1, pp. 105--153, 2022.

\bibitem{zhang2017correction}
S.~Zhang and Z.~Wang, ``Correction: inferring passenger denial behavior of taxi drivers from large-scale taxi traces,'' \emph{Plos one}, vol.~12, no.~2, p. e0171876, 2017.

\bibitem{chen2024micro}
L.~Chen, Y.~Zhang, M.~Liu, and H.~Wang, ``Micro-macro spatial-temporal graph-based encoder-decoder for trajectory recovery,'' \emph{arXiv preprint arXiv:2404.19141}, 2024.

\bibitem{berte2024enhancing}
\BIBentryALTinterwordspacing
M.~Bert{\`e}, R.~Ibrahimli, L.~Koopmans, P.~Valga{\~n}{\'o}n, N.~Zomer, and D.~Colombi, ``Enhancing stop location detection for incomplete urban mobility datasets,'' \emph{arXiv preprint arXiv:2407.11579}, 2024. [Online]. Available: \url{https://arxiv.org/abs/2407.11579}
\BIBentrySTDinterwordspacing

\bibitem{pang2024finding}
J.~Pang, M.~A. Sabir, Z.~Wang, A.~Hu, X.~Yang, H.~Yu, and Q.~Huang, ``Finding a taxi with illegal driver substitution activity via behavior modelings,'' \emph{IEEE Transactions on Intelligent Transportation Systems}, 2024.

\bibitem{pang2018learning}
J.~Pang, J.~Huang, Y.~Du, H.~Yu, Q.~Huang, and B.~Yin, ``Learning to predict bus arrival time from heterogeneous measurements via recurrent neural network,'' \emph{IEEE Transactions on Intelligent Transportation Systems}, vol.~20, no.~9, pp. 3283--3293, 2018.

\bibitem{kumar2017visual}
D.~Kumar, J.~Bezdek, S.~Rajasegarar, C.~Leckie, and M.~Palaniswami, ``A visual-numeric approach to clustering and anomaly detection for trajectory data,'' \emph{The Visual Computer}, vol.~33, pp. 265--281, 2017.

\bibitem{chen2011real}
C.~Chen, D.~Zhang, P.~Samuel~Castro, N.~Li, L.~Sun, and S.~Li, ``Real-time detection of anomalous taxi trajectories from gps traces,'' in \emph{International Conference on Mobile and Ubiquitous Systems: Computing, Networking, and Services}.\hskip 1em plus 0.5em minus 0.4em\relax Springer, 2011, pp. 63--74.

\bibitem{boateng2023gps}
C.~Boateng, K.~Yang, S.~Ghoreishi, J.~Jang, M.~Jan, J.~Conniff, B.~Furht, S.~Moshfeghi, D.~Newman, R.~Tappen \emph{et~al.}, ``Abnormal driving detection using gps data,'' in \emph{IEEE 20th International Conference on Smart Communities: Improving Quality of Life using AI, Robotics and IoT (HONET)}, 2023.

\bibitem{park2024unsupervised}
J.~Park and M.~Kim, ``An unsupervised learning framework for urban driving anomaly detection using gps trajectories,'' \emph{IEEE Transactions on Intelligent Transportation Systems}, vol.~25, no.~3, pp. 3341--3353, 2024.

\bibitem{yu2022deep}
W.~Yu and Q.~Huang, ``A deep encoder-decoder network for anomaly detection in driving trajectory behavior under spatio-temporal context,'' \emph{International Journal of Applied Earth Observation and Geoinformation}, vol. 115, p. 103115, 2022.

\bibitem{li2023temporal}
Y.~Li, S.~Zhou, and M.~Wang, ``Temporal attention-based trajectory representation learning for anomaly detection in public transport,'' \emph{IEEE Access}, vol.~11, pp. 78\,533--78\,545, 2023.

\bibitem{zhang2023semi}
L.~Zhang, Y.~Dong, H.~Farah, A.~Zgonnikov, and B.~van Arem, ``Data-driven semi-supervised machine learning with surrogate safety measures for abnormal driving behavior detection,'' \emph{arXiv preprint arXiv:2312.04610}, 2023.

\bibitem{yang2021driving}
Y.~Yang, J.~Yan, J.~Guo, Y.~Kuang, M.~Yin, S.~Wang, and C.~Ma, ``Driving behavior analysis of city buses based on real-time gnss traces and road information,'' \emph{Sensors}, vol.~21, no.~12, p. 4018, 2021.

\bibitem{ozdemir2018hybrid}
E.~Ozdemir, A.~E. Topcu, and M.~K. Ozdemir, ``A hybrid hmm model for travel path inference with sparse gps samples,'' \emph{Transportation}, vol.~45, pp. 233--246, 2018.

\bibitem{zhang2015shortest}
\BIBentryALTinterwordspacing
D.~Zhang, T.~He, and Y.~Lin, ``Shortest path and vehicle trajectory aided map-matching for low-frequency gps data,'' \emph{Transportation Research Part C: Emerging Technologies}, vol.~55, pp. 328--339, 2015. [Online]. Available: \url{https://doi.org/10.1016/j.trc.2015.01.003}
\BIBentrySTDinterwordspacing

\bibitem{chen2017road}
\BIBentryALTinterwordspacing
C.~Chen, Y.~Li, and J.~Yu, ``Road traffic anomaly detection via collaborative path inference from gps snippets,'' \emph{Sensors}, vol.~17, no.~3, p. 550, 2017. [Online]. Available: \url{https://doi.org/10.3390/s17030550}
\BIBentrySTDinterwordspacing

\bibitem{zheng2024generative}
Y.~Zheng, H.~Koh, M.~Jin, L.~Chi, K.~T. Phan, S.~Pan, Y.-P. Chen, and W.~Xiang, ``Generative semi-supervised graph anomaly detection,'' \emph{arXiv preprint arXiv:2402.11887}, 2024.

\bibitem{song2024novel}
W.~Song, X.~Li, P.~Chen, J.~Chen, J.~Ren, and Y.~Xia, ``A novel graph structure learning based semi-supervised framework for anomaly identification in fluctuating iot environment,'' \emph{Computer Modeling in Engineering \& Sciences}, vol. 140, no.~3, pp. 3001--3016, 2024.

\bibitem{chen2025semi}
J.~Chen, S.~Fu, Z.~Ma, M.~Feng, T.~Wirjanto, and Q.~Peng, ``Semi-supervised anomaly detection with extremely limited labels in dynamic graphs,'' \emph{arXiv preprint arXiv:2501.15035}, 2025.

\bibitem{tian2023sad}
S.~Tian, J.~Dong, J.~Li, W.~Zhao, X.~Xu, B.~Wang, B.~Song, C.~Meng, T.~Zhang, and L.~Chen, ``Sad: Semi-supervised anomaly detection on dynamic graphs,'' \emph{arXiv preprint arXiv:2305.13573}, 2023.

\bibitem{latif2023detecting}
H.~Latif-Martínez, J.~Suárez-Varela, A.~Cabellos-Aparicio, and P.~Barlet-Ros, ``Detecting contextual network anomalies with graph neural networks,'' \emph{arXiv preprint arXiv:2312.06342}, 2023.

\bibitem{zheng2023correlation}
Y.~Zheng, H.~Koh, M.~Jin, L.~Chi, K.~T. Phan, S.~Pan, Y.-P. Chen, and W.~Xiang, ``Correlation-aware spatial-temporal graph learning for multivariate time-series anomaly detection,'' \emph{arXiv preprint arXiv:2307.08390}, 2023.

\bibitem{etemad2019trajectory}
M.~Etemad, A.~S. J{\'u}nior, A.~Hoseyni, J.~Rose, and S.~Matwin, ``A trajectory segmentation algorithm based on interpolation-based change detection strategies.'' in \emph{EDBT/ICDT Workshops}, 2019, p.~58.

\bibitem{guo2018gps}
S.~Guo, X.~Li, W.-K. Ching, R.~Dan, W.-K. Li, and Z.~Zhang, ``Gps trajectory data segmentation based on probabilistic logic,'' \emph{International Journal of Approximate Reasoning}, vol. 103, pp. 227--247, 2018.

\end{thebibliography}
\endgroup

\end{document}